\documentclass[runningheads]{llncs}

\usepackage[mobile]{eccv}

\usepackage{eccvabbrv}

\usepackage{graphicx}
\usepackage{booktabs}
\usepackage{multirow}
\usepackage{xspace}
\usepackage[accsupp]{axessibility}  
\usepackage{tikz}
\newcommand{\fcirc}[1]{%
  \tikz[baseline=(C.base)]\node[%
    circle, fill=black, inner sep=0.6pt, minimum size=1.55ex
  ] (C) {\textcolor{white}{\scriptsize #1}};%
}
\usepackage[most]{tcolorbox}

\newcommand{\method}{\textsc{PER}\xspace}
\newcommand{\methodlong}{Patch Embedding Replacement\xspace}

\usepackage{hyperref}

\usepackage{orcidlink}

\begin{document}

\title{Look But Don't Touch with Sparse Autoencoders for Unlearning in Diffusion Models} 
\titlerunning{Look But Don't Touch}

\author{Enrico Cassano\inst{1}\orcidlink{0009-0004-6490-6503} \and
Riccardo Renzulli\inst{1}\orcidlink{0000-0003-0532-5966} \and
Rayyan Ahmed\inst{2}\orcidlink{0009-0008-0678-5319} \and
Marco Grangetto\inst{1}\orcidlink{0000-0002-2709-7864} \and
Stephan Alaniz\inst{2}\orcidlink{0000-0003-3541-2163}}

\authorrunning{E.~Cassano et al.}

\institute{University of Turin, Computer Science Department, Italy \\
\email{\{name.surname\}@unito.it}\\
 \and
LTCI, Télécom Paris, Institut Polytechnique de Paris, France \\
\email{\{name.surname\}@telecom-paris.fr}}

\maketitle

\begin{abstract} 
Sparse autoencoders (SAEs) have recently been proposed as interpretable tools for concept-level manipulation, under the assumption that isolated features can serve as controllable intervention points. In this work, we systematically evaluate this assumption in the context of object erasure and steering in diffusion models. 
We show that while SAEs reliably detect and localize semantic concepts within diffusion model activations, direct intervention in their latent space frequently induces out-of-distribution activations, resulting in severe visual artifacts. 
To disentangle detection from intervention, we use SAE activations purely as semantic detectors to identify image regions containing the target object, and replace those patch embeddings with the ones that do not contain it. This detection-based replacement preserves the diffusion model’s activation statistics and produces significantly cleaner erasure results than latent steering.
Our findings reveal a fundamental gap between concept detection and concept intervention in diffusion models: monosemantic or sparse features are not inherently suitable as control knobs for steering. These results position SAEs as powerful interpretability tools for analyzing generative models, but highlight important limitations when used for direct manipulation, such as unlearning. We release the code at \url{https://eidoslab.github.io/PER/}.
\end{abstract}    
\section{Introduction}
\label{sec:intro}
\begin{figure}[t]
    \centering
    \includegraphics[width=1\linewidth]{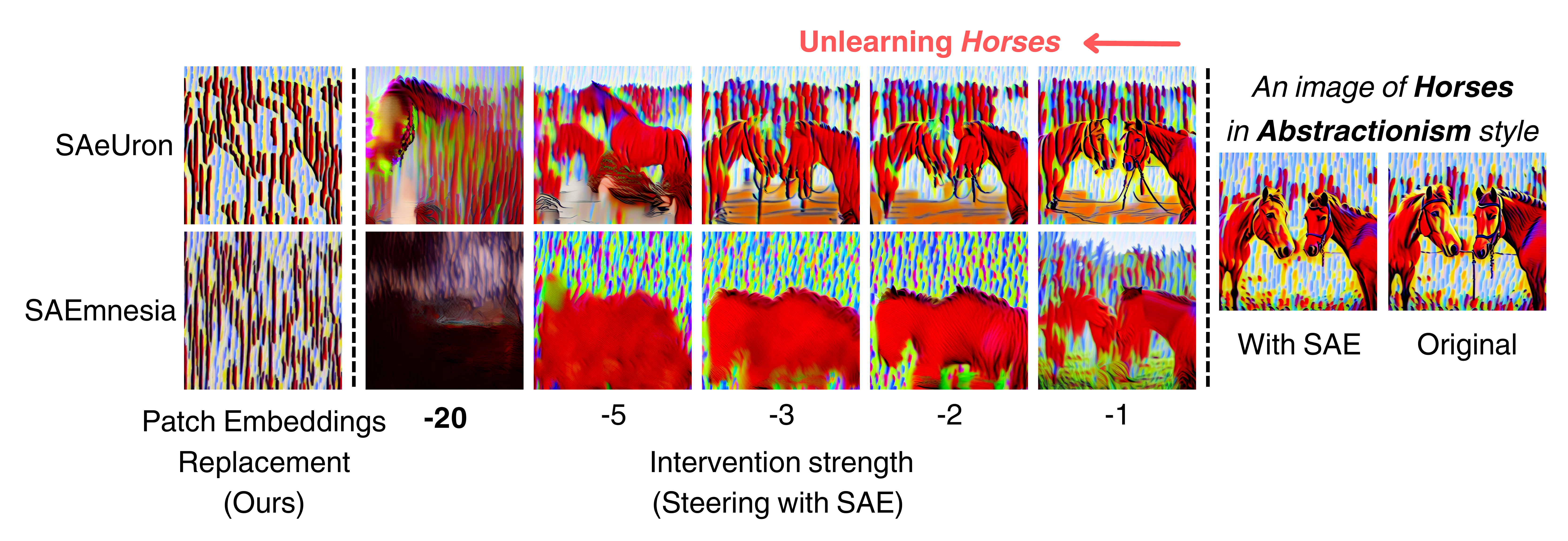}
    \caption{\textbf{Effect of SAE-based activation steering under varying intervention strengths and feature patch replacement when unlearning ``Horses''}. Both SAeUron (top) and SAEmnesia (bottom) exhibit severe visual artifacts at large negative multipliers, while weaker interventions fail to fully erase the concept. Multipliers highlighted in bold are the values selected by each method for horse unlearning.}
    \label{fig:teaser}
\end{figure}

Controlling content generation in text-to-image models~\cite{rombach2022high} has become increasingly critical as these systems are deployed at scale. When models produce explicit content, copyrighted works, or other undesirable outputs, the ability to selectively erase specific concepts, ensuring the model can no longer represent or generate them, becomes essential~\cite{mu_survey}. A core challenge in machine unlearning is identifying where and how concepts are represented inside diffusion models (DMs), particularly given the phenomenon of \textit{polysemanticity}, where each neuron can encode multiple unrelated concepts simultaneously~\cite{bricken2023monosemanticity}. Sparse Autoencoders (SAEs) decompose dense neural activations into sparse, \textit{monosemantic} latents that activate almost exclusively for specific concepts and serve as the fundamental units of neural network representations~\cite{bereska2024mechanistic}. 
Recent work has explored SAEs as a tool for interpretable concept erasure~\cite{cywinski2025saeuron, cassano2026saemnesia} and editing~\cite{kim2025concept, surkov2024one, tinaz2025emergence} in DMs. These methods all follow a similar pipeline: (i) encode DM activations through the SAE encoder and (ii) steer activations with the SAE latents that represent the concepts to erase or edit. This approach offers interpretability advantages, since we can precisely identify which latents represent which concepts. However, we can observe in \cref{fig:teaser} that steering with SAEs produces systematic artifacts when unlearning the concept ``Horses'' from Stable Diffusion (SD) v1.5~\cite{rombach2022high} fine-tuned on UnlearnCanvas~\cite{zhang2024unlearncanvas}, which is a stylized image dataset for benchmarking unlearning methods in DMs. In fact, concept erasure via SAE reconstruction frequently produces activations that lie outside the DM’s in-distribution (ID) manifold, resulting in visual artifacts.
Nonetheless, SAE-based approaches excel at detecting the presence of a concept, an important step for unlearning, and thus present a promising alternative to traditional unlearning methods based on fine-tuning~\cite{gandikota2023erasing,zhang2024forget,kumari2023ablating} or closed-form weight modifications~\cite{gandikota2024unified}. While prior work has shown that SAEs excel at detection but underperform at steering in language models~\cite{wu2025axbench}, we hereby systematically characterize a concrete failure mode in SAE-based unlearning for DMs and provide a concrete alternative. We show that using SAEs only for concept detection rather than direct manipulation avoids out-of-distribution (OOD) activations. For each representation of the DM containing the unwanted concept, we replace the concept-containing embeddings with concept-free ones drawn from the same feature map. By doing so, we remove unwanted concepts while maintaining visual coherence. Moreover, every activation in our processed output originates from the DM's forward pass, guaranteeing that all activations are ID within the DM.
Our key contributions can be summarized as follows: \fcirc{1} We show that SAE reconstruction under latent intervention produces activations outside the DM's manifold (\cref{sec:ood-artifacts}). \fcirc{2} We introduce a detection-based framework for SAE-guided concept removal through patch embeddings replacement (PER), avoiding the distributional mismatch (\cref{sec:per}). \fcirc{3} Our method improves the quality of the unlearned images, eliminates the costly grid search for steering intervention strength and scores up to an average of 95.33\% on UnlearnCanvas, which is an improvement of 3.82\% over state-of-the-art for SAE-based object unlearning (\cref{sec:results}). We also achieve promising results in NSFW removal and adversarial attacks.

\section{Related work}\label{sec:relatedwork}
\noindent\textbf{Concept erasure in DMs.}
Machine unlearning aims to remove the influence of particular data or concepts from a trained model without necessitating retraining, ensuring that the resulting model works as if the target information were never encountered during training~\cite{mu_survey}. 
We follow the established convention in which unlearning is framed as concept erasure~\cite{feng2025surveygenmu}, aiming to prevent the generation of specific concepts while maintaining overall generation quality and preserving unrelated concepts. The distributed and entangled representation of concepts in DMs makes this problem especially challenging. Recent work involving full model fine-tuning to unlearn concepts has explored various approaches: ESD~\cite{gandikota2023erasing} and CA~\cite{kumari2023ablating} eliminate anchor concepts through fine-tuning with negative guidance; EDiff~\cite{wu2024erasediff} formulates data forgetting as a constrained optimization problem; SA~\cite{heng2023selective} substitutes unwanted data distributions with surrogate distributions. Fine-tuning-free methods include SalUn~\cite{fan2024salun} and SHS~\cite{wu2024scissorhands}, which identify parameters to modify using saliency maps or connection sensitivity; FMN~\cite{zhang2024forget}, which introduces a re-steering loss applied exclusively to attention layers; SPM~\cite{lyu2024one}, which incorporates linear adapters to directly prevent unwanted content propagation; SEOT~\cite{li2024get}, which eliminates unwanted content from text embeddings; UCE~\cite{gandikota2024unified}, which modifies cross-attention weights via closed-form solutions. In contrast to these approaches, our method leverages SAEs to achieve interpretable concept erasure at inference time without modifying model weights, ensuring that activations remain within the DM's training distribution.

\noindent\textbf{Erasing concepts with SAEs in DMs.} In recent years, the exploitation of SAEs for preventing the generation of specific concepts has become an emerging topic. Kim \etal~\cite{kim2025concept} leverage SAEs in the latent space of text embeddings to precisely steer the generation away from a given concept. Cywi\'{n}ski and Deja~\cite{cywinski2025saeuron} proposed SAeUron, which applies unsupervised SAEs after cross-attention layer in UNet-based DMs. While SAeUron achieves promising results when unlearning styles in UnlearnCanvas, its effectiveness when unlearning objects is lower. With unsupervised SAEs, concepts can still be distributed across many neurons. Therefore, Cassano \etal~\cite{cassano2026saemnesia} introduced SAEmnesia, which trains SAEs with supervision to achieve higher feature centralization, binding each concept to a single, interpretable neuron. Similarly, H{\"a}rle \etal~\cite{harle2025measuring} proposed Guided Sparse Autoencoders (G-SAEs), a method that conditions latent representations on labeled concepts to better understand and influence the internal dynamics of large language models (LLMs). In this work, we adapt G-SAEs to DMs. Surkov \etal~\cite{surkov2024one} use SAEs trained SDXL Turbo in its 1-step setting for editing images, but they struggle in the delete object task. In contrast to these works, our method uses SAEs exclusively for concept detection, identifying patches containing unwanted concepts and replacing them with concept-free alternatives from the same feature map, thereby ensuring that all activations remain within the DM's training distribution while avoiding the distributional mismatch inherent in reconstruction-based approaches.

\noindent{\textbf{Limitations of SAE-based steering.}} While SAEs excel at decomposing neural activations into interpretable features, recent work reveals systematic limitations when using these features for steering. Wu \etal~\cite{wu2025axbench} introduce a comprehensive benchmark evaluating SAE-based steering against simple baselines across concept detection and model steering tasks. Their findings show how SAEs significantly underperform even basic methods and fail to approach the effectiveness of standard prompting or fine-tuning. They demonstrate that supervised dictionary learning methods consistently outperform unsupervised SAEs for both detection and intervention tasks. O'Brien \etal~\cite{o2024steering} show that editing SAE latent values to control refusal behavior in language models improves safety metrics but causes widespread performance degradation on unrelated benchmarks. Critically, they find this degradation occurs regardless of which features are manipulated, suggesting fundamental limitations in latents-based steering. Mayne \etal~\cite{mayne2024can} investigate why SAE decompositions of steering vectors fail to preserve steering properties, identifying two core issues: steering vectors exhibit distributional properties incompatible with SAE training assumptions, and meaningful interventions require negative feature coefficients which SAEs cannot represent. Building on this prior evidence of SAE-based steering in language models, we identify an analogous limitation in DMs unlearning. To our knowledge, we are the first to systematically characterize this failure mode in SAE-based interventions for DMs, wherein steering via SAE features generates OOD activations that degrade both visual fidelity and unlearning effectiveness. We show that our methodology, PER, avoids this distributional drift by preserving the model's learned activation manifold throughout the unlearning.    
\section{Unlearning with SAEs}\label{sec:method}
\begin{figure}[t]
    \centering
    \includegraphics[width=0.85\linewidth]{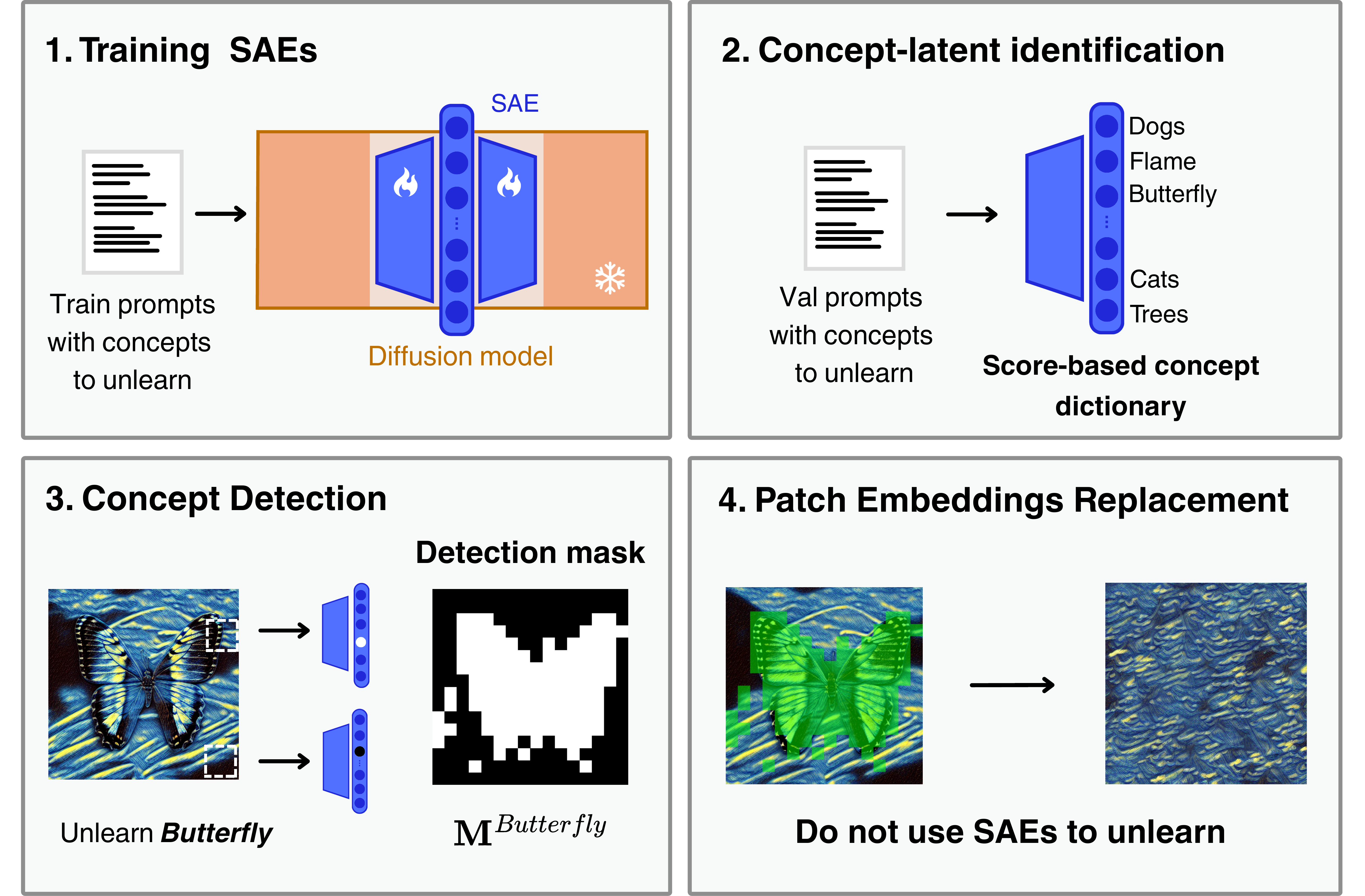}
    \caption{\textbf{Overview of the proposed pipeline.}
First, SAEs are trained on DM activations using prompts containing the concepts to be removed. Next, a score-based analysis identifies the SAE latents associated with each concept, forming a concept–latent dictionary. During inference, these latents are used to detect concept-containing patches, producing a spatial detection mask. Finally, instead of steering with SAEs, detected patch embeddings are replaced with embeddings sampled from non-detected locations, removing the concept while keeping activations in-distribution.}
    \label{fig:pipeline}
\end{figure}

\cref{fig:pipeline} provides an overview of our approach. Instead of steering with SAE latents, we replace detected concept-containing patch embeddings with concept-free ones sampled from the same feature map, this removes the concept while keeping activations ID.
\subsection{Training SAEs in DMs}
In this work, we apply SAEs to the residual stream of cross-attention layers on the denoising network of DMs. At timestep $t$, we extract output feature maps $\mathbf{X}_t \in \mathbb{R}^{H \times W \times D}$, where $H$ and $W$ are the height and width of the feature map, and $D$ is the hidden dimension. Each spatial position within the feature map corresponds to a patch in the input image. We denote by $\mathbf{x} \in \mathbb{R}^D$ a generic patch embedding vector given as input to the SAE, which is composed of an encoder and a decoder structure. The encoder expands the dimension $D$ to dimension $N$ by a fixed factor. The encoding and decoding operations are formulated as~\cite{bricken2023monosemanticity}:
\begin{equation}
    \mathbf{v} = \text{ReLU}(\mathbf{W}_{\text{enc}}(\mathbf{x}-\mathbf{b}_\text{pre}) + \mathbf{b}_\text{enc}), \quad \hat{\mathbf{x}} = \mathbf{W}_\text{dec}\mathbf{v} + \mathbf{b}_\text{pre},
\end{equation}
where $\mathbf{v}$ denotes sparse latent codes, $\hat{\mathbf{x}}$ represents the reconstructed input, and weight matrices $\mathbf{W}_\text{enc} \in \mathbb{R}^{N \times D}$ and $\mathbf{W}_\text{dec} \in \mathbb{R}^{D \times N}$ parameterize the encoder and decoder, while $\mathbf{b}_\text{pre} \in \mathbb{R}^D$ and $\mathbf{b}_\text{enc} \in \mathbb{R}^N$ are learnable bias terms. Instead of vanilla ReLU SAEs, we train TopK SAEs~\cite{Makhzani2013kSparseA} that enforce sparsity by retaining only the $k$ most strongly activated latents. This mechanism zeroes out all but the top $k$ pre-activation values, maintaining the dominant features while guaranteeing a fixed sparsity level:
\begin{equation}
    \mathbf{z} = \text{TopK}(\mathbf{v}), \quad \hat{\mathbf{x}} = \mathbf{W}_\text{dec}\mathbf{z} + \mathbf{b}_\text{pre}.
\end{equation}
The vector $\mathbf{v}$ contains pre-TopK activations, while $\mathbf{z}$ holds the sparsified representation after TopK. Training TopK SAEs requires balancing reconstruction fidelity against latent usage. For a batch of $B$ samples, the objective is:
\begin{equation}
    \label{eq:sae_loss}
    \mathcal{L}_\text{unsupSAE} = \frac{1}{B} \sum_{b=1}^{B}\big\|\mathbf{x}^{(b)} - \hat{\mathbf{x}}^{(b)}\big\|_2^2+ \alpha\,\mathcal{L}_\text{aux},
\end{equation}
where the first term measures reconstruction quality and $\mathcal{L}_\text{aux}$~\cite{cywinski2025saeuron} addresses dead latents, \ie features that rarely activate during training,  weighted by scalar $\alpha$. Additionally, when employing SAEmnesia and G-SAEs, we also compute the corresponding proposed supervised losses to enforce concept-latent mapping.

\subsection{Concept-latent identification}\label{sec:concept-latent-identification} To identify which SAE latents correspond to a given concept, we follow Cywi\'{n}ski and Deja~\cite{cywinski2025saeuron} and use a score function that measures the correspondence between a latent (i.e. neuron) and a concept. Given a dataset of activations $\mathcal{D} = \mathcal{D}_c \cup \mathcal{D}_{\neg c}$, where $\mathcal{D}_c$ are the activations containing the concept $c$ and $\mathcal{D}_{\neg c}$ are the activations without it, the score for latent $n$ at denoising timestep $t$ is:
\begin{equation}
\label{eq:score}
        \text{score}(n, t, c, \mathcal{D}) = \frac{\mu(n, t, \mathcal{D}_c)}{\sum_{j=1}^N \mu(j, t, \mathcal{D}_c) + \delta} 
        - \frac{\mu(n, t, \mathcal{D}_{\neg c})}{\sum_{j=1}^N \mu(j, t, \mathcal{D}_{\neg c}) + \delta}
\end{equation}
where $\delta$ is a small constant to prevent division by zero and $\mu(n, t, \mathcal{D}) = \frac{1}{|\mathcal{D}|}\sum_{x \in \mathcal{D}} z_n$ denotes the mean activation of the $n$-th latent over dataset $\mathcal{D}$ at timestep $t$ ($t$ is omitted from $z_n$ for brevity). Intuitively, this score ranks all latents for a given concept $c$: latents with high scores activate strongly in the presence of $c$ and weakly otherwise. We denote the set of top-scoring latents for concept $c$ as:
\begin{equation}
    \mathcal{F}_c = \{ n  : n \text{ is a top-scoring latent selected for concept } c \}
\end{equation}
The size $|\mathcal{F}_c|$ depends on the underlying pipeline.

\subsection{Concept detection}\label{sec:concept-detection}
Let $\mathbf{Z} \in \mathbb{R}^{H \times W \times N}$ denote the SAE latent activations for all spatial locations, where $Z_{h,w,n}$ is the activation of latent $n$ at position $(h,w)$. 
For a concept $c$, associated with a set of latents $\mathcal{F}_c$, we define a binary detection mask $\mathbf{M}^c \in \{0,1\}^{H \times W}$ as
\begin{equation}
M^c_{h,w}
=
\mathbf{1}\!\left(
\exists\, n \in \mathcal{F}_c :
Z_{h,w,n} > \mu(n,t,\mathcal{D})
\right),
\end{equation}
where $\mu(n,t,\mathcal{D})$ denotes the average activation of latent $n$ at timestep $t$ over the validation set $\mathcal{D}$, and $\mathbf{1}(\cdot)$ is the indicator function. 
A spatial location is marked as concept-containing if at least one concept-associated latent exceeds its dataset mean activation. We employ this detection mask when steering G-SAEs~\cite{harle2025measuring} and SAEs on SDXL Turbo~\cite{surkov2024one} and with PER.   

\subsection{Steering with SAEs}
SAE-based steering modifies the contribution of concept-associated latents and re-injects the modified activation into the denoising network. We adopt two suppression strategies based on the SAE. The first strategy performs a direct additive displacement in activation space. As mentioned before, for G-SAEs and SAEs trained on SDXL Turbo, we modify their original steering formulation with the binary detection mask as
\begin{equation}
\tilde{\mathbf{X}}_{h,w} = 
\mathbf{X}_{h,w} - A_{h,w}\mathbf{W}_\text{dec}[:, c]
\end{equation}
where $\mathbf{A} = \gamma_c\mathbf{M}^c$ and $\gamma_c$ is a negative multiplier, namely the intervention strength. For simplicity, we omit the timestep $t$ in the notation and use $c$ as the index of the decoder column responsible for concept $c$.
This formulation applies a linear perturbation aligned with the learned feature directions. SAeUron and SAEmnesia employ an alternative suppression strategy that acts directly on $\mathbf{Z}$, replacing the activation with its SAE reconstruction.
We define a latent-level gating mask as

\begin{equation}
G_{h,w,n}
=
\mathbf{1}\!\left(
n \in \mathcal{F}_c
\;\wedge\;
Z_{h,w,n} > \mu(n,t,\mathcal{D})
\right).
\end{equation}
Selected latents are scaled by a negative multiplier $\gamma_c$,
normalized by the average activation on concept samples
$\mu(n,t,\mathcal{D}_c)$:
\begin{equation}
Z'_{h,w,n}
=
(1 - G_{h,w,n})\, Z_{h,w,n}
+
G_{h,w,n}\,
\gamma_c \, \mu(n,t,\mathcal{D}_c)\, Z_{h,w,n}.
\end{equation}
The steered activation is then reconstructed through the decoder:
\begin{equation}
\tilde{\mathbf{X}}_{h,w}
=
\mathbf{W}_{\mathrm{dec}}\, \mathbf{Z}'_{h,w,:}
+
\mathbf{b}_{\mathrm{pre}},
\end{equation}
and replaces the original activation at the spatial position $(h,w)$. In both cases, steering relies on the decoder dictionary to define semantically meaningful directions in representation space. The key difference lies in whether the intervention applies a localized displacement directly to the denoising network's original activation or steers it in the SAE latent space.

\subsection{\methodlong (\method)}\label{sec:per}
 We observed that current SAE-based unlearning steering techniques exhibit degenerate failure modes characterized by: 1) the generation of OOD activations, and 2) systematic visual artifacts resulting from these perturbations. We analyze these effects in detail in \cref{sec:ood-artifacts}. Moreover, prior approaches require a grid search over the intervention strength $\gamma_c$ for each concept $c$, introducing additional complexity and instability. In contrast, we proposed a simple \methodlong (\method) method that removes multiplier-based latent manipulation.
 Given the binary detection mask $\mathbf{M}^c \in \{0,1\}^{H \times W}$ defined in \cref{sec:concept-detection}, we erase concept $c$ by replacing detected patch embeddings at locations $(h,w)$ where $\mathbf{M}^c_{h,w}=1$ with activations drawn from non-detected locations:
\begin{equation}
\label{eq:patch_concept}
\begin{aligned}
    \tilde{\mathbf{X}}_{h,w} &= (1 - \mathbf{M}^c_{h,w})\mathbf{X}_{h,w} + \mathbf{M}^c_{h,w}\mathbf{X}_{h',w'}\\
    \text{where} \quad (h',w') &\sim \mathcal{U}(\{(i, j) \mid \mathbf{M}^c_{i,j} = 0\}).
\end{aligned}
\end{equation}



Here, the spatial location $(h', w')$ is uniformly sampled from positions not activating the concept.
Since concept information may propagate to neighboring patches through spatial correlations, we optionally dilate the detection mask using a padding factor $p \geq 0$. Specifically, for each detected location, all patches within a spatial distance $p$ are also marked as concept-containing, yielding a dilated mask $\mathbf{M}^c(p)$. When $p=0$, no dilation is applied.
In the rare case where all spatial locations are detected (\ie, $\mathbf{M}^c = \mathbf{1}$), the method falls back to Gaussian noise replacement. Other replacement strategies are reported in Appendix~\ref{sec:appendix_variations}. Importantly, all replacement activations originate from the same forward pass of the denoising network. As a result, $\tilde{\mathbf{X}}$ remains within the model's native activation distribution, avoiding the distributional shift and the corresponding visual artifacts introduced by multiplier-based latent interventions.

\section{Experiments and Results}
\label{sec:results}
Our experiments reveal the OOD behavior of existing SAE unlearning approaches, resulting in generated artifacts. Furthermore, we show that \method compares favorably across multiple dimensions: computational efficiency, unlearning effectiveness on the UnlearnCanvas benchmark and adversarial robustness. We apply \method on top of four existing SAE-based pipelines, SAeUron~\cite{cywinski2025saeuron}, SAEmnesia~\cite{cassano2026saemnesia}, G-SAE~\cite{harle2025measuring} and SAE trained on SDXL Turbo~\cite{surkov2024one}. For a fair comparison, we compare \method in the experimental setups of each pipeline.

\subsection{Experimental setup}
\label{sec:experiments}

\noindent\textbf{SAEs and DMs.} Although our pipeline includes SAE training as first step, we can directly adopt the pre-trained SAE models released by previous pipelines, when available. Note that, to the best of our knowledge, G-SAEs have not been applied to DMs unlearning before our work. We evaluate \method across different DM architectures: SD v1.5~\cite{rombach2022high}, which serves as the primary testbed given that most existing erasure techniques are developed and benchmarked on this architecture, and the more recent SDXL Turbo, assessing the generality of our approach. For SD v1.5, we use the SAE on UNet block \texttt{up.1.1}, consistent with prior work~\cite{cywinski2025saeuron, cassano2026saemnesia}, while for SDXL Turbo we use block \texttt{up.0.1}~\cite{surkov2024one}. 

\noindent\textbf{Dataset and evaluation metrics.}
Our primary evaluation uses the UnlearnCanvas benchmark~\cite{zhang2024unlearncanvas}, which consists of 20 object classes and 50 artistic styles, with prompts of the form \texttt{An image of \{\textit{object}\} in \{\textit{style}\} style}. Vision Transformer-based classifiers are employed to measure three key metrics: 1) Unlearning Accuracy (UA), which quantifies the proportion of samples from target concept prompts that are misclassified (\ie successful unlearning); 2) In-domain Retain Accuracy (IRA), measuring classification accuracy on retained concepts within the same domain; 3) Cross-domain Retain Accuracy (CRA), assessing accuracy on concepts from different domains. To quantify visual artifacts, we employ Qwen2-VL-7B-Instruct~\cite{Qwen2VL} as an automated evaluator to measure the Artifacts Rate (AR) across generated images on which unlearning was applied. Comparisons with other unlearning methodologies can be found in~\cref{sec:baselines_uc} of the Appendix, and additional experiments performed for NSFW removal on the I2P benchmark~\cite{schramowski2023safe} on SD v1.4 are reported in~\cref{sec:appendix_nudity} of the Appendix.


\subsection{The brittleness of SAE-based Unlearning}\label{sec:ood-artifacts}

We find that current SAE-based unlearning techniques suffer from degenerate failure modes and identify two primary issues: the generation of OOD activations and the resulting systematic visual artifacts.



\begin{figure}[h]
      \centering
      \includegraphics[width=0.95\linewidth]{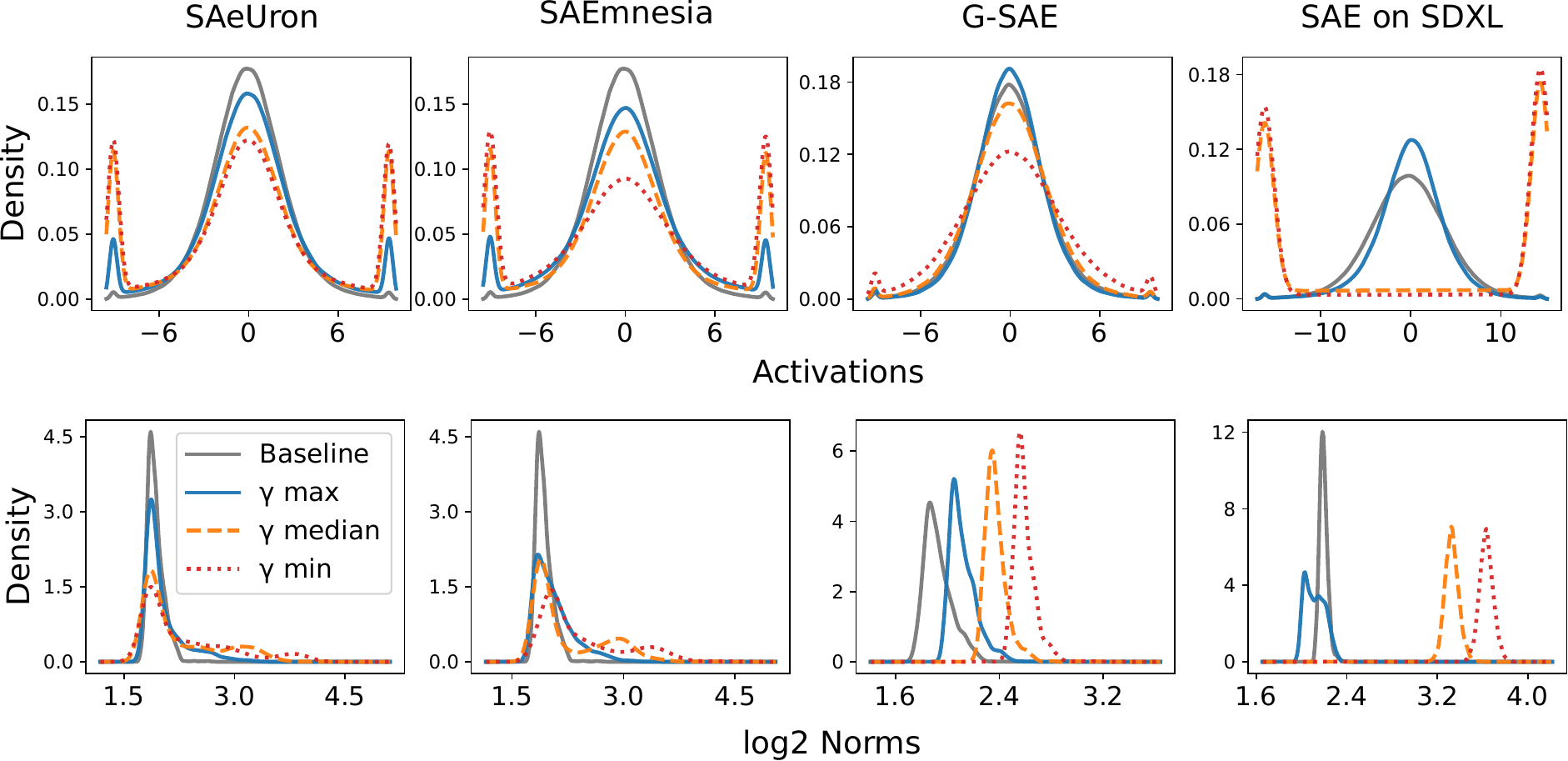}
     \caption{\textbf{Distributions at the hooked layer comparing original and SAE-steered activations across four models (SAeUron, SAEmnesia, GSAE, SDXL Turbo)}. Top row: Activation distributions; bottom row: L2 norm distributions (log scale). $\gamma$ max, $\gamma$ median and $\gamma$ min are the maximum, median and minimum values, respectively, of the $\gamma$ ranges of the presented methods.}
     \label{fig:sae_distribution}
\end{figure}

\noindent\textbf{Activation distribution analysis.} To quantify the effect of SAE-based unlearning through negative multipliers, we collect patch-level activations with and without intervention. Following the setting of UnlearnCanvas, we generate one image for every object class $c$ using the template \texttt{An image of \{\textit{$c$}\}} and record the activations for every patch embedding at the SAE layer 
where the SAE detects the concept $c$.
Further, we repeat the process and record the embeddings after SAE unlearning of concept $c$ as described in \cref{sec:method}. Since the negative multiplier $\gamma_c$ is typically different for every concept $c$, we collect embeddings for the minimum, median and maximum value of $\gamma$ on UnlearnCanvas to capture the range of the manipulated embedding space distributions. We plot the per-dimension activation distributions (\cref{fig:sae_distribution} top) and find that using the median negative multiplier produces a wider and flatter activation profile, with 27.6\%, 28\%, 18.8\%, and 83.7\% of values falling outside the original distributions for SAeUron, SAEmnesia, G-SAE, and SDXL Turbo respectively. 
This distribution shift is particularly extreme for SDXL Turbo where half activations lie outside the original activation distribution. These results show that multiplier-based unlearning pushes reconstructed activations outside the DM's training distribution, especially with larger multipliers, rather than performing a targeted removal that preserves the activation manifold. In \cref{fig:sae_distribution} (bottom), we show the L2 norm distributions, revealing that, while original norms are concentrated around $10^{2}$, steered activations exhibit a heavy tail reaching $10^{3}$. We see that 51.3\%, 44.7\%, 99.5\%, and 100\% of activations generated with the median negative multipliers do not overlap with the norm distribution for SAeUron, SAEmnesia, G-SAE, and SDXL Turbo respectively.
Both of these results support our claim that multiplier-based unlearning pushes the reconstructed activations outside the activation distribtion of the DM.

\begin{figure}[h]
    \centering
    \includegraphics[width=1\linewidth]{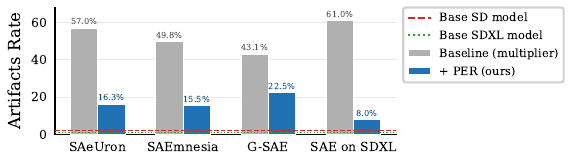}
    \caption{\textbf{AR (\%) for SAeUron, SAEmnesia, G-SAE, and SAE on SDXL Turbo on UnlearnCanvas.} Applying \method to different base SAEs consistently improves AR.}
    \label{fig:artifacts_rate.pdf}
\end{figure}

\noindent\textbf{Artifacts Rate.}
\label{sec:AR}
A consequence of the OOD activations from multiplier-based interventions is the presence of artifacts in the generated images. 
We use Qwen2-VL-7B-Instruct to measure the AR across images generated with and without unlearning. It is prompted to detect artifacts in outputs of SAE-based unlearning on UnlearnCanvas concepts (employed prompt reported in~\cref{sec:appendix_artifacts_rate} of the Appendix). We define the AR as the percentage of images that the VLM marks as containing artifacts out of the total images submitted. As shown in~\cref{fig:artifacts_rate.pdf}, when no unlearning is applied, the AR is 2\% for SD 1.5 and 1\% for SDXL Turbo, both computed over 1020 generated images. 
Multiplier-based methods have higher ARs: SAeUron produces artifacts in 57\% of generated images, SAEmnesia in 49.8\%, G-SAE in 43.1\%, and the SAE trained on SDXL Turbo in 61\%. When we apply \method, the AR is lower across all baselines: from 57\% to 16.3\% for SAeUron, from 49.8\% to 15.5\% for SAEmnesia, from 43.1\% to 22.5\% for G-SAE and from 61\% to 8\% for SAE on SDXL Turbo. These results support our hypothesis: replacing concept-containing patches with in-distribution patches mitigates the distributional mismatch (we analyze the distribution of the replacement patches in~\cref{sec:appendix_patch_ood} of the Appendix) and artifacts. \cref{fig:infra_class_artifacts} provides a per-concept breakdown for SAeUron (the same analysis performed for SAEmnesia is reported in~\cref{sec:appendix_per_concept_saemnesia} of the Appendix). 
\begin{figure}[t]
    \centering
    \includegraphics[width=1\linewidth]{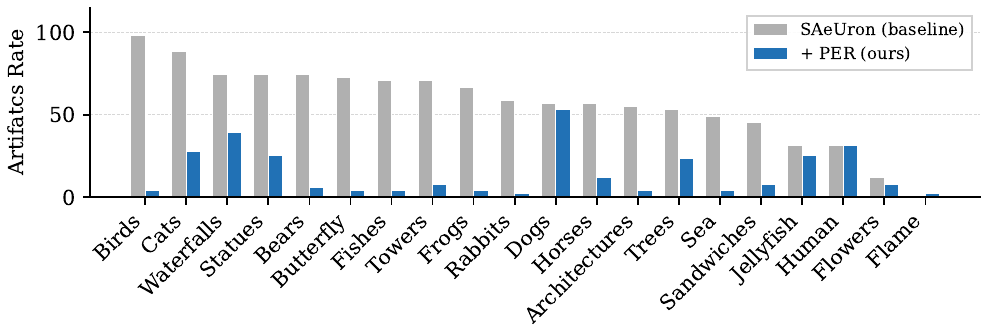}
    \caption{\textbf{Per-concept AR (\%) on UnlearnCanvas for SAeUron with and without \method.} Concepts are sorted by baseline AR (descending). \method reduces the overall AR from 57.0\% to 16.3\%, with the largest
    improvements on concepts where intervention strength is most disruptive (\eg, ``Birds'', ``Cats'' and ``Bears'').
    }
    \label{fig:infra_class_artifacts}
\end{figure}
Concepts with high baseline AR improve the most: ``Birds'' drops from 98.0\% to 3.9\%, ``Cats'' from 88.2\% to 27.5\%, and ``Bears'' from 74.5\% to 5.9\%. Notably, ``Dogs'' and ``Humans'' show limited or no improvement (56.9\%\,$\to$\,52.9\% and 31.4\%\,$\to$\,31.4\%). 

\noindent\textbf{Human Evaluation on AR.}
\label{app:human_evaluation}
We conducted two user studies to validate our AR results. The first compares the AR measured by Qwen2-VL-7B-Instruct against humans on a 150-image subset. As reported in ~\cref{tab:ar_three_sources}, the 18 annotators reproduce the ordering AR(SD) $<$ AR(+PER) $\ll$ AR(SAEmnesia) with Krippendorff's $\alpha = 0.76$, denoting substantial agreement. 
\begin{table}[h]
\centering
\begin{minipage}{0.55\columnwidth}
\centering
\begin{tabular}{l c c c }
    \hline
            & \textbf{Base SD} & \textbf{SAEmnesia} & \textbf{+PER} \\
    \hline
    Humans  & \phantom{0}0.0 & 90.7 & 10.0 \\
    VLM     & 10.0           & 39.5 & 14.0 \\
    \hline
\end{tabular}
\end{minipage}%
\hfill
\begin{minipage}{0.42\columnwidth}
\caption{AR (\%, $\downarrow$ better) on a 150-image subset: human majority vote (18 annotators) vs.\ Qwen2-VL.}
\label{tab:ar_three_sources}
\end{minipage}
\end{table}
This confirms that the VLM-AR in~\cref{fig:artifacts_rate.pdf} is a conservative lower bound on PER's improvements. The second study collects pairwise preferences between SAEmnesia and +PER outputs on 47 evaluation pairs among 26 retained annotators. The results are reported in~\cref{tab:pairwise_pref}. Out of 1183 valid responses, 927 were non-tie (exactly one of SAEmnesia or +PER was preferred) and 256 were ties. Among the non-tie responses, +PER is preferred over SAEmnesia in 885 of 927 cases, or 95.5\%; this margin is highly significant ($p<10^{-4}$), with a 95\% confidence interval ranging from 93.9 to 96.6\%. Ties between SAEmnesia and +PER account for 21.6\% of responses, of which 231 were rated ``both bad'' and only 25 ``both good'', indicating that tied pairs reflect joint failure rather than joint success. SAEmnesia is preferred on only 2 pairs, both in the Architectures concept.

\begin{table*}[t]
\centering
\caption{Pairwise human preference between +PER and SAEmnesia. We report the results for 26 retained annotators across 47 evaluation pairs (1{,}183 valid judgments). ``\% of total'' is over all judgments; ``\% of group'' is within each split. Among non-tie judgments, +PER is preferred in 95.5\% of cases.}
\label{tab:pairwise_pref}
\setlength{\tabcolsep}{6pt}
\begin{tabular}{llccc}
    \hline
    \textbf{Split} & \textbf{Outcome} & \textbf{Judgments} & \textbf{\% of total} & \textbf{\% of group} \\
    \hline
    Non-tie & \textbf{+PER preferred} & \textbf{885} & \textbf{74.8} & \textbf{95.5} \\
            & SAEmnesia preferred     & 42  & 3.6  & 4.5   \\
            & \emph{Subtotal}         & 927 & 78.4 & 100.0 \\
    \hline
    Tie     & Both bad                & 231 & 19.5 & 90.2 \\
            & Both good               & 25  & 2.1  & 9.8   \\
            & \emph{Subtotal}         & 256 & 21.6 & 100.0 \\
    \hline
    \multicolumn{2}{l}{\textbf{Total}} & \textbf{1183} & \textbf{100.0} & --- \\
\end{tabular}
\end{table*}


\subsection{Quantitative results}
\label{sec:quantitative_results}


\noindent\textbf{Computational efficiency.} \method can be applied to any SAE-based unlearning pipeline, requiring no model fine-tuning, no additional SAE training, and no $\gamma_c$ search. In contrast, existing methods require a grid search on a validation set over $\gamma_c$ values per concept: SAeUron, SAEmnesia, G-SAE, and SAE trained on SDXL Turbo must evaluate 7, 7, 6, and 9 values of $\gamma$, respectively (ranges are reported in~\cref{sec:appendix_gammas}). \method removes this costly search across all methods.

\noindent\textbf{UnlearnCanvas.} \cref{tab:core_results} presents the performance of \method on the UnlearnCanvas benchmark for object unlearning. The base metrics reported by UnlearnCanvas do not account for the visual quality of generated images when unlearning concepts. 
\begin{table}[h]
\small
\centering
\caption{\textbf{Evaluation metrics (\%) on object concept unlearning using the UnlearnCanvas benchmark.} Comparing different SAE-based unlearning methods to \method with the same SAE. The best result for each metric is highlighted in bold.}
\label{tab:core_results}
\begin{tabular}{lcccccc}
\textbf{Method} & \textbf{UA} (↑) & \textbf{IRA} (↑) & \textbf{CRA} (↑) & \textbf{Avg.} (↑) & \textbf{AR} ($\downarrow$) & \textbf{GA} (↑)\\
\hline
SAeUron & 87.16 & 85.57 & 74.14 & 82.29 & 57.0 & 72.47 \\
+ \method & 85.37 & 81.14 & 86.55 & \textbf{84.35} & \textbf{16.3} & \textbf{84.19} \\
\hline
SAEmnesia & 94.65 & 91.39 & 88.48 & 91.51 & 49.8 & 81.18 \\
+ \method & 91.37 & 91.92 & 97.97 & \textbf{93.45} & \textbf{15.5} &  \textbf{91.44} \\
\hline
\textsc{G-SAE} & 78.14 & 96.14 & 95.56 & 89.94 & 43.1 & 81.69 \\
+ \method & 94.02 & 96.11 & 95.87 & \textbf{95.33} & \textbf{22.5} & \textbf{90.88}\\
\hline
SDXL Turbo SAE & 13.92 & 85.55 & - & 49.74 & 61.0 & 46.15 \\
+ \method  & 14.41 & 86.06 & - & \textbf{50.24} & \textbf{8.0} & \textbf{64.16}\\
\end{tabular}
\end{table}
In fact, a method could achieve high unlearning scores while producing images with high AR. To capture this dimension, we incorporate $(100 - \text{AR})$ in the average score computation and name it Global Average (GA). This ensures that the final aggregate metric rewards both effective unlearning \emph{and} the ability to generate clean, high-quality images.
When applied to SAeUron, \method reaches GA of 84.19\%, a 11.72\% improvement over the multiplier-based baseline. Notably, the best improvement is on the CRA metric, where our method scores 86.55\% compared to 74.14\%. When applied to SAEmnesia, \method achieves 91.44\% GA score, improving over the 81.18\% baseline by 10.26\%. We again achieve substantial gains on CRA (97.97\% vs.\ 88.48\%) while keeping IRA slightly above the baseline (91.92\% vs.\ 91.39\%). The consistent CRA improvements across both pipelines confirm that replacing concept-containing patches with in-distribution activations preserves unrelated concepts more effectively than multiplier-based intervention. \method also significantly improves the GA when applied to G-SAE, yielding a 9.19\% gain. In this case, the highest gain is on the UA metric, mainly due to the already high CRA.
For SAE trained on SDXL Turbo, the CRA could not be computed fairly since, unlike the other models, SDXL Turbo was not fine-tuned on UnlearnCanvas. We still evaluated the unlearning on objects, showing a slight improvement in UA, IRA, and AR, which brings the GA from 46.15\% to 64.16\%.
Additional results with different padding values are in \cref{sec:appendix_unlearncanvas}.

\begin{figure}[t]
    \centering
    \begin{minipage}[c]{0.53\textwidth}
        \centering
        \includegraphics[width=\linewidth]{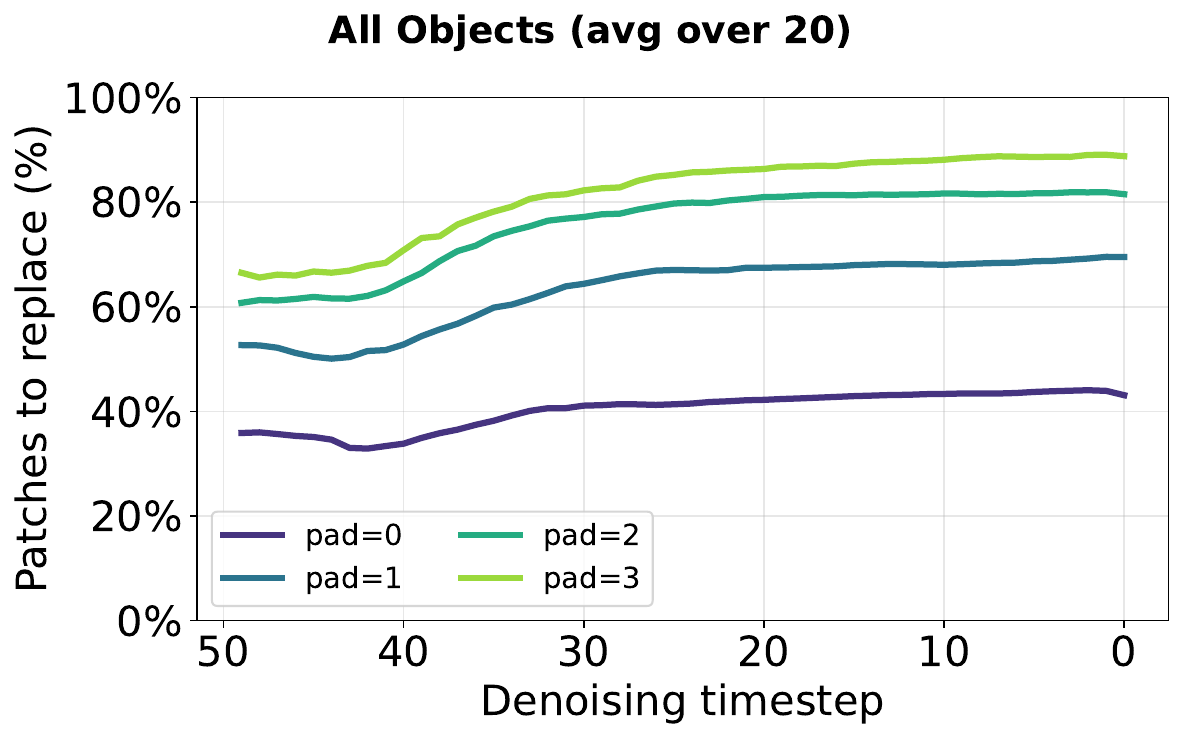}
    \end{minipage}
    \hfill
    \begin{minipage}[c]{0.43\textwidth}
        \caption{Ratio of SAEmnesia patches replaced across timesteps for padding values $p \in \{0, 1, 2, 3\}$.}
        \label{fig:padding_saemnesia}
    \end{minipage}
\end{figure}

\noindent\textbf{Adversarial Attacks.} We follow the experimental setup of SAeUron and SAEmnesia. By applying adversarial attacks to \method, we go beyond surface-level unlearning comparisons and provide a more complete picture of our approach's behavior. We follow the UnlearnDiffAtk benchmark~\cite{zhang2024generate} on the objects of UnlearnCanvas, optimizing 5-token adversarial prefixes for 40 iterations with a learning rate of 0.01. Results are reported in \cref{tab:adv_attk}. 
\begin{table}[h]
\centering
\caption{\textbf{Adversarial robustness (UnlearnDiffAtk).} Object UA (\%) before and after attack, and attack effectiveness.}
\label{tab:adv_attk}
\begin{tabular}{llccc}
\hline
\textbf{Pipeline} & \textbf{Method} & \textbf{Before ($\uparrow$)} & \textbf{After ($\uparrow$)} & \textbf{Eff. ($\downarrow$)} \\
\hline
SAeUron & Baseline & 83.70 & 34.20 & 49.50 \\
        & + \method & 84.60 & 28.30 & 56.30 \\
\hline
SAEmnesia & Baseline & 97.60 & 57.50 & 40.10 \\
        & + \method & 91.10 & 56.20 & \textbf{34.90} \\
\hline
\end{tabular}
\end{table}
When applied to the SAeUron pipeline, \method achieves comparable pre-attack UA (84.60\%) but shows a larger drop after the adversarial attack, with an effectiveness value of 56.30 compared to 49.50 for the baseline. When applied to the SAEmnesia pipeline, the trend reverses. \method achieves the lowest attack effectiveness at 34.90, improving over the SAEmnesia baseline of 40.10. While the pre-attack UA is lower (91.10\% vs.\ 97.60\%), the post-attack UA remains competitive (56.20\% vs.\ 57.50\%), resulting in a smaller overall degradation. Additional results are reported in~\cref{sec:appendix_adv}.
\noindent\textbf{Padding Analysis.} We first motivate the choice of $p=1$ through the experimental results reported in~\cref{sec:appendix_unlearncanvas}. Setting $p=0$ leads to a clear drop in unlearning effectiveness: with SAeUron, $p=0$ yields a UA of only 49.12\% compared to 85.37\% at $p=1$, a drop of over 36 pp, despite achieving higher CRA (97.95\% vs.\ 86.55\%), suggesting that an overly sparse intervention fails to suppress the target concept reliably. Padding beyond $p=1$ does not consistently improve results: while $p=2$ achieves a slightly higher UA under SAeUron (88.82\% vs.\ 85.37\%), it underperforms $p=1$ on the overall average under SAEmnesia (91.70\% vs.\ 93.45\%), and also yields a worse adversarial robustness effectiveness score (38.60\% vs.\ 34.90\%). This suggests that $p=1$ is the most reliable choice. \cref{fig:padding_saemnesia} further illustrates this trade-off by showing the fraction of patches marked for replacement at each denoising timestep for increasing padding factors. At $p = 0$, only the patches directly identified as concept-containing are replaced, resulting in a sparse intervention concentrated at later timesteps. As $p$ increases, a larger portion of the feature map is marked. 
\subsection{Qualitative results}
\begin{figure}[h]
    \centering
    \includegraphics[width=0.9\linewidth]{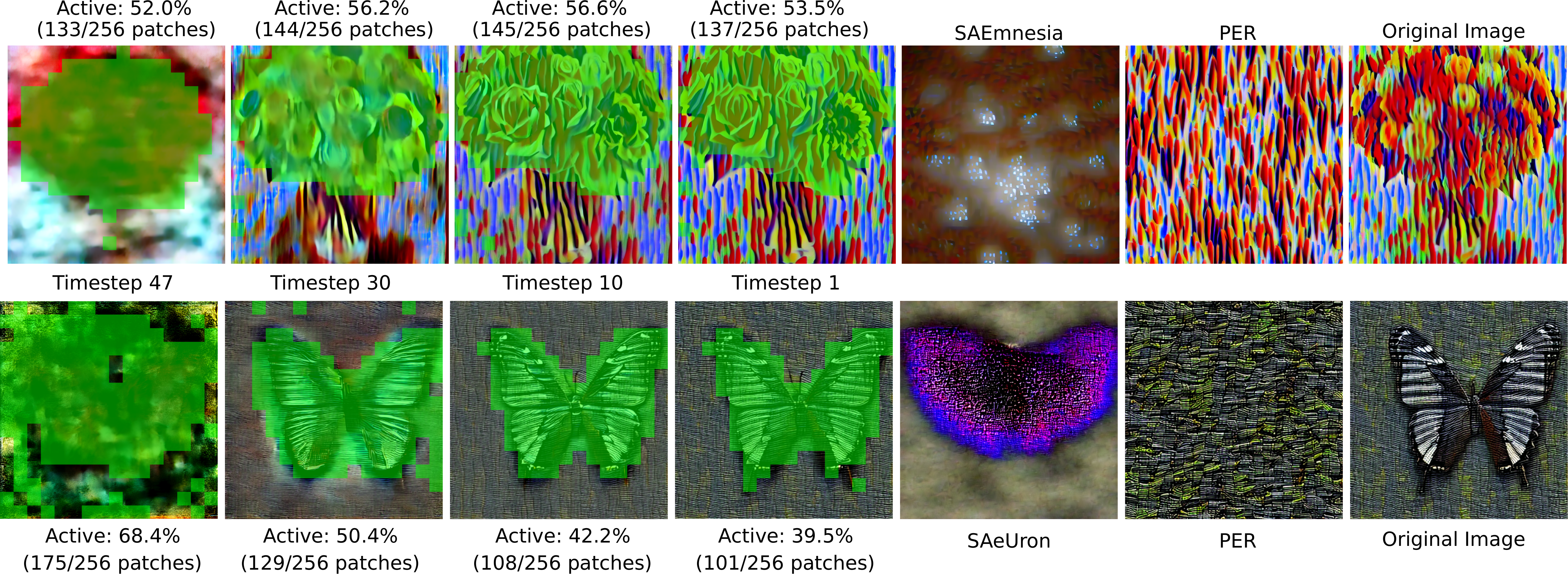}
   \caption{Concept detection masks ($p=0$) overlaid on intermediate timesteps for SAeUron and SAEmnesia. As generation progresses, the mask tightens around the objects.}
    \label{fig:qual_saemnesia}
\end{figure}

\begin{figure}[h]
    \centering
    \includegraphics[width=0.9\linewidth]{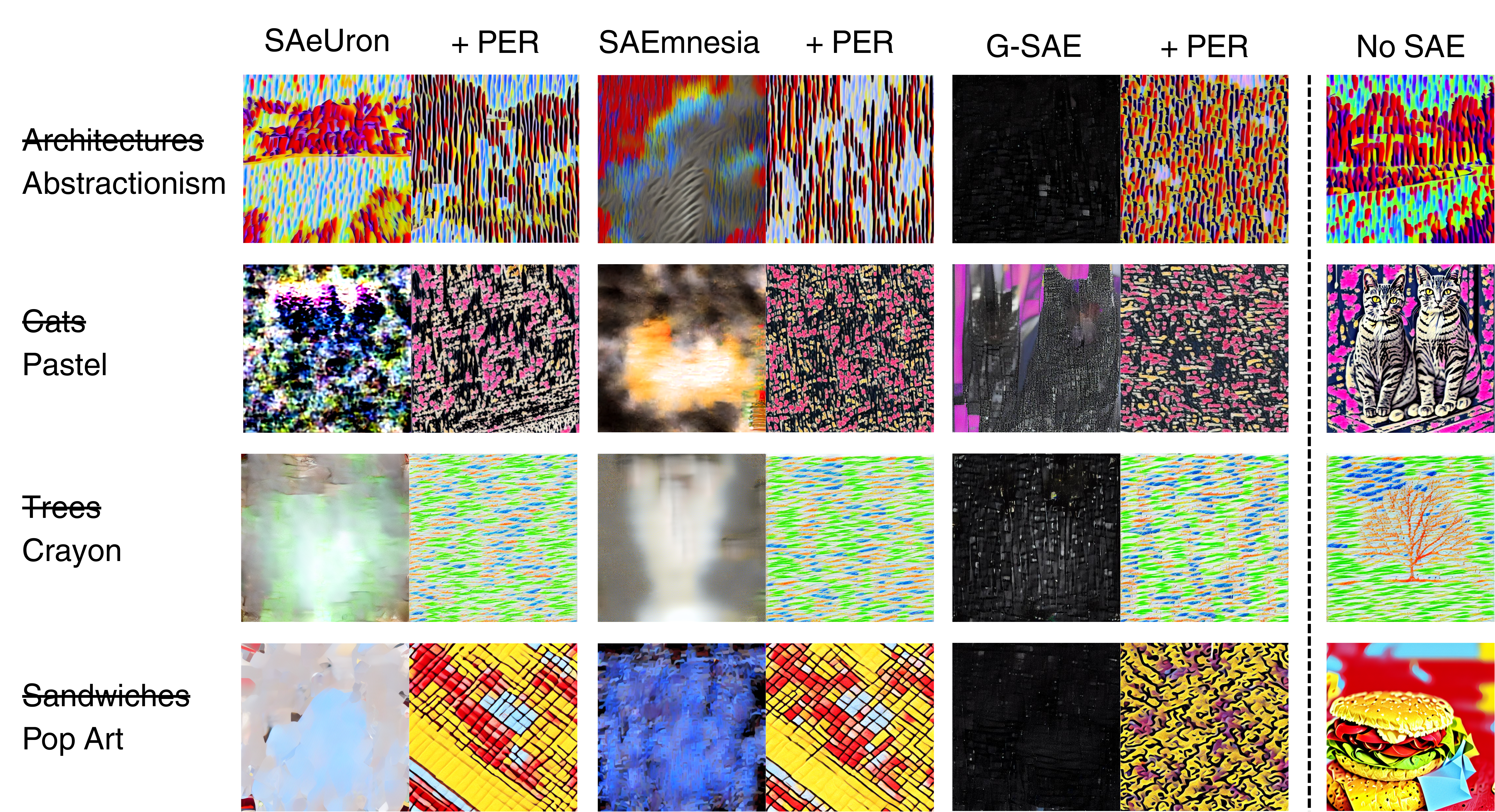}
   \caption{\textbf{Qualitative comparison of SAE-based unlearning methods with and without PER.} Existing SAE-based steering methods can produce visual artifacts or corrupted images. In contrast, PER removes the object while preserving coherent style, demonstrating that detection-based patch replacement produces cleaner results.}
    \label{fig:other_qual_saemnesia}
\end{figure}

\cref{fig:qual_saemnesia} shows the patches detected as concept containing by the SAE latents alone ($p=0$), overlaid on the feature maps at selected timesteps. The green regions correspond to patches where the Eq.~\ref{eq:patch_concept} is satisfied. We can see that SAE latents localize the target concept with remarkable precision. 
At early denoising timesteps (high $t$), the feature map is still noisy and the detected patches are more scattered; as denoising progresses toward $t=1$, the concept becomes increasingly localized and the mask sharpens around the target object. Concept information in a contiguous feature map does not have hard spatial boundaries: the influence of a concept naturally bleeds into neighboring patches, which the $p=0$ mask leaves unmarked. Expanding the detected region by a padding factor $p > 0$ is therefore a natural choice to increase coverage. In \cref{fig:other_qual_saemnesia}, we observe that unlearning with \method produces artifact-free results, preserving original style and improving visual quality.

\section{Conclusion}
\label{sec:conclusion}
We studied SAEs for unlearning in DMs and identified a concrete failure mode. While SAE latents can reliably detect concepts, directly steering them with multiplier-based interventions often produces OOD activations, leading to artifacts. We introduced PER, a detection-based framework that leverages SAEs only for concept localization, avoiding latent manipulation. By replacing concept-containing patch embeddings with ID activations of the same feature map, PER preserves the DM’s activation statistics while removing the target concept. Experiments across multiple SAEs show that PER consistently reduces artifacts, improves cross-domain retention, and achieves higher performance on UnlearnCanvas, NSFW removal, and adversarial attacks without intervention strength tuning. Our work shows the importance of separating detection from intervention when designing interpretable control mechanisms for generative models.
\section*{Acknowledgements}
We acknowledge the CINECA award under the ISCRA initiative, for the availability of high performance computing resources and support. The `Mechanistically-Grounded Adaptive AI' action has received funding from the
European Union, via the oc4-2025-TES-02 issued and implemented by the ENFIELD
project, under the grant agreement No 101120657.
This work is also supported by Hi! PARIS and ANR/France 2030 program (ANR-23-IACL-0005).

%
%
\bibliographystyle{splncs04}
\bibliography{main}

\newpage
\appendix
\clearpage
\section{Appendix}

\subsection{Baselines for UnlearnCanvas.}
\label{sec:baselines_uc}
Table~\ref{tab:uc_core_results} reports the performance of the state-of-the-art methods on object concept unlearning for the Unlearn Canvas benchmark. \method applied to the \textsc{G-SAE} pipeline outperforms all the compared methods.

\begin{table}[!ht]
\small
\centering
\caption{\textbf{State-of-the-art methods on object concept unlearning tested on the UnlearnCanvas benchmark.} The best result is highlighted in bold.}
\label{tab:uc_core_results}
\begin{tabular}{lcccc}
\toprule
\textbf{Method} & \textbf{UA} (↑) & \textbf{IRA} (↑) & \textbf{CRA} (↑) & \textbf{Avg.} (↑)\\
\midrule
ESD & 92.15 & 55.78 & 44.23 & 64.05 \\
FMN & 45.64 & 90.63 & 73.46 & 69.91 \\
UCE & 94.31 & 39.35 & 34.67 & 56.11 \\
CA & 46.67 & 90.11 & 81.97 & 72.92 \\
SalUn & 86.91 & 96.35 & 99.59 & 94.28 \\
SEOT & 23.25 & 95.57 & 82.71 & 67.18 \\
SPM & 71.25 & 90.79 & 81.65 & 81.23 \\
EDiff & 86.67 & 94.03 & 48.48 & 76.39 \\
SHS & 80.73 & 81.15 & 67.99 & 76.62 \\
SAeUron & 87.16 & 85.57 & 74.14 & 82.29 \\
SAEmnesia & 94.65 & 91.39 & 88.48 & 91.51 \\
\textsc{G-SAE} & 78.14 & 96.14 & 95.56 & 89.94 \\
\midrule
SAeUron + \method (p=1) & 85.37 & 81.14 & 86.55 & 84.35 \\
SAEmnesia + \method (p=1) & 91.37 & 91.92 & 97.97 & 93.45 \\
\textsc{G-SAE} + \method (p=1) & 94.02 & 96.11 & 95.87 & \textbf{95.33} \\
\bottomrule
\end{tabular}
\end{table}

\subsection{Artifacts Rate.}
\label{sec:appendix_artifacts_rate}
The prompt used with the VLM to quantify the AR in the generated images is reported in Figure~\ref{fig:prompt}.

\begin{figure}[h]
\centering
\begin{tcolorbox}[title={Qwen2-VL-7B-Instruct prompt}]
\textbf{Prompt:}
This image was generated by a diffusion model. Look at the overall visual style and texture of the image. An image is BAD if ANY of the following are true: 
1. It contains a large distinct region (such as a gray, white, or black blob/patch) that clearly breaks from the surrounding style or texture. 
2. It is mostly a single uniform color (gray, black, white) with little to no meaningful visual content, texture, or style.
3. It contains a faint ghostly shape or silhouette against a flat background.
An image is GOOD if it has rich visual content with a consistent style throughout,
even if abstract, noisy, or heavily stylized.
Answer only GOOD or BAD.
\end{tcolorbox}

\caption{AR prompt for Qwen2-VL-7B-Instruct.}
\label{fig:prompt}
\end{figure}

\subsection{NSFW unlearning.}
\label{sec:appendix_nudity}

\begin{table}
\centering
\caption{\textbf{NSFW unlearning evaluation on the I2P benchmark.}
Columns: Armpits (Arm), Belly (Bel), Buttocks (But), Feet (Ft),
Female Breasts/Genitalia (FB/FG), Male Breasts/Genitalia (MB/MG).
For each SAE-based method, \method is applied on top and shares the same CLIP/FID scores.}
\begin{tabular}{lrrrrrrrrrrr}
\hline
Method & Arm & Bel & But & Ft & FB & FG & MB & MG & Total & CLIP$\uparrow$ & FID$\downarrow$ \\
\hline
FMN & 43 & 117 & 12 & 59 & 155 & 17 & 19 & 2 & 424 & 30.39 & 13.52 \\
CA & 153 & 180 & 45 & 66 & 298 & 22 & 67 & 7 & 838 & 31.37 & 16.25 \\
AdvUn & 8 & 0 & 0 & 13 & 1 & 1 & 0 & 0 & 28 & 28.14 & 17.18 \\
Receler & 48 & 32 & 3 & 35 & 20 & 0 & 17 & 5 & 160 & 30.49 & 15.32 \\
MACE & 17 & 19 & 2 & 39 & 16 & 0 & 9 & 7 & 111 & 29.41 & 13.42 \\
CPE & 10 & 8 & 2 & 8 & 6 & 1 & 3 & 2 & 40 & 31.19 & 13.89 \\
UCE & 29 & 62 & 7 & 29 & 35 & 5 & 11 & 4 & 182 & 30.85 & 14.07 \\
SLD-M & 47 & 72 & 3 & 21 & 39 & 1 & 26 & 3 & 212 & 30.90 & 16.34 \\
ESD-x & 59 & 73 & 12 & 39 & 100 & 6 & 18 & 8 & 315 & 30.69 & 14.41 \\
ESD-u & 32 & 30 & 2 & 19 & 27 & 3 & 8 & 2 & 123 & 30.21 & 15.10 \\
\hline
SAeUron & 7 & 1 & 3 & 2 & 4 & 0 & 0 & 1 & 18 & 30.89 & 14.37 \\
\quad + \method & 2 & 7 & 0 & 3 & 9 & 0 & 0 & 0 & 21 & 30.89 & 14.37 \\
\hline
SAEmnesia & 7 & 17 & 2 & 5 & 11 & 2 & 2 & 1 & 47 & 30.98 & 14.72 \\
\quad + \method & 7 & 11 & 0 & 2 & 7 & 0 & 0 & 0 & 27 & 30.98 & 14.72 \\
\hline
\textsc{SAEmnesia-top2} & 1 & 3 & 1 & 0 & 4 & 0 & 0 & 0 & 9 & 30.98 & 14.72 \\
\quad + \method top2 & 2 & 2 & 0 & 0 & 2 & 1 & 0 & 0 & 7 & 30.98 & 14.72 \\
\hline
SD v1.4 & 148 & 170 & 29 & 63 & 266 & 18 & 42 & 7 & 743 & 31.34 & 14.04 \\
SD v2.1 & 105 & 159 & 17 & 60 & 177 & 9 & 57 & 2 & 586 & 31.53 & 14.87 \\
\hline
\end{tabular}
\label{tab:nudity_merged}
\end{table}

\begin{figure}[t]
    \centering
\includegraphics[width=0.9\linewidth]{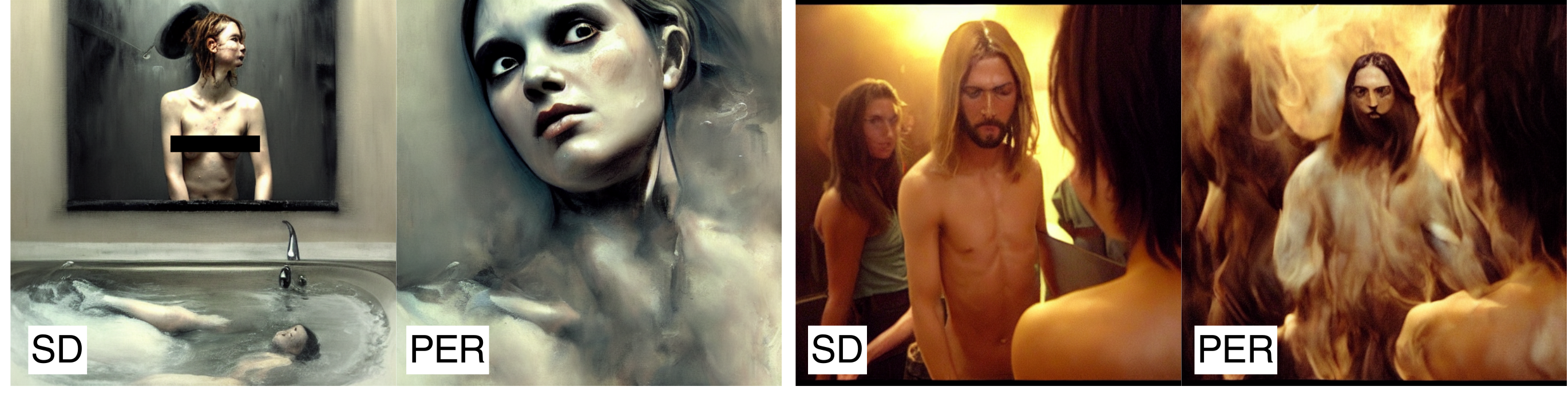}
    \caption{Nudity removal qualitative results (I2P NSFW).}
    \label{fig:nudity}
\end{figure}

\begin{figure}
    \centering
    \includegraphics[width=.7\linewidth]{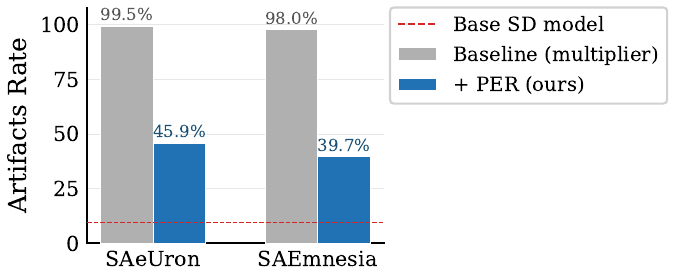}
    \caption{\textbf{AR(\%) for SAeUron and SAEmnesia nudity unlearning.} As per UnlearnCanvas, we have a consistent improvement when applying \method.}
    \label{fig:ar_nudity}
\end{figure}

For NSFW unlearning, we count the number of exposed body parts detected across generated images following the established methodology in literature~\cite{cywinski2025saeuron, cassano2026saemnesia}. \cref{tab:nudity_merged} reports the number of exposed body parts detected across 4703 generated images for all evaluated methods, including the application of \method on the SAeUron and SAEmnesia pipelines. The unmodified Stable Diffusion v1.4 model produces 743 total detections, serving as the upper bound reference.
When applied to the SAeUron pipeline, \method yields 21 total detections compared to 18 for the baseline. Although patch substitution does not reduce the detection count in this setting, the method still provides gains in image quality as per~\cref{fig:ar_nudity}, and eliminates the need for multiplier search.
When applied to the SAEmnesia pipeline with a single top-1 latent, \method reduces the total count from 47 to 27, a 42.5\% reduction. The improvement is consistent across categories, with the complete elimination of buttocks, female genitalia, male breasts, and male genitalia detections. The most persistent categories are belly (11) and armpits (7), which remain challenging due to their frequent co-occurrence with non-nudity content. The improvement is also present in the AR of the unlearnt images, which drops from 98\% to 39.7\% when applying \method.
We also apply \method to the top-2 latent variant of SAEmnesia. In this setting, the baseline already achieves 9 detections, and \method further reduces this to 7. The narrower gap suggests that the additional latent already captures most of the residual NSFW content that patch substitution would otherwise address.
These results indicate that \method provides more effective NSFW removal than multiplier-based intervention while remaining competitive with state-of-the-art unlearning methods shown in the upper portion of \cref{tab:nudity_merged}. For the CLIP score and FIDs we follow the methodology proposed in the SAE-based unlearning literature~\cite{cywinski2025saeuron, cassano2026saemnesia}. In addition to the quantitative results, in~\cref{fig:nudity} we also report a qualitative example for NSFW removal. On the contrary of Unlearn Canvas, in which DMs are finetuned per style, the NSFW test is performed on photorealistic images, showing the transferability of our methodology beyond narrow-scoped benchmarks.

\subsection{\method Variations.}
\label{sec:appendix_variations}
Table~\ref{tab:results_appendix} reports ablation results comparing two patch substitution
strategies on the SAEmnesia pipeline, for $p \in \{1, 2\}$. An intuitive design
choice for the replacement strategy is to substitute each concept-containing patch with its
nearest concept-free neighbor in the feature map, as spatially adjacent patches are likely
to share similar visual content and thus produce a coherent output. 
However, as shown in Table~\ref{tab:results_appendix}, Random substitution (as done in \method) consistently outperforms this alternative strategy across all metrics and both padding values, suggesting that introducing diversity in the replacement source is beneficial for unlearning effectiveness. Replacing marked patches with Gaussian noise at $p=0$ yields a competitive but slightly lower average than Random (89.86\% vs.\ 90.95\%), further confirming that same-image substitution is the stronger strategy.

\begin{table}[h]
\setlength{\tabcolsep}{5pt}
\small
\centering
\caption{Ablation over patch substitution strategies for $p \in \{1, 2\}$ on the
SAEmnesia pipeline. Random (R): each marked patch is replaced by a patch
sampled uniformly at random from the concept-free patches in the feature map.
Nearest (N): each marked patch is replaced by its spatially closest concept-free
neighbor. Despite the intuitive appeal of spatial proximity, Random substitution
outperforms Nearest across all metrics and both padding values.}
\begin{tabular}{cccccc}
\hline
\textbf{Padding} & \textbf{Method} & \textbf{UA} & \textbf{IRA} & \textbf{CRA} & \textbf{Avg} \\
\hline
1 & R & 77.84\% & 95.73\% & 99.28\% & 90.95\% \\
1 & N & 59.51\% & 96.25\% & 99.45\% & 85.07\% \\
\hline
2 & R & 86.18\% & 86.71\% & 98.30\% & 90.40\% \\
2 & N & 71.18\% & 91.69\% & 99.09\% & 87.32\% \\
\hline
\end{tabular}
\label{tab:results_appendix}
\end{table}

\subsection{Additional Experiments on Adversarial Attacks.}
\label{sec:appendix_adv}

Table~\ref{tab:adv_attk_appendix} extends the adversarial robustness results of Section~\ref{sec:results} by reporting results for both $p=1$ and $p=2$. The two padding values yield comparable robustness across both pipelines. For SAEmnesia, $p=1$ achieves the lowest attack effectiveness (34.90), while for SAeUron both values are similar (56.30 vs.\ 55.70).

\begin{table}[h]
\centering
\caption{\textbf{Adversarial robustness (UnlearnDiffAtk).} Object UA (\%) before and after attack, and attack effectiveness. We report the results for $p=1$ and $p=2$.}
\label{tab:adv_attk_appendix}
\begin{tabular}{llccc}
\hline
\textbf{Pipeline} & \textbf{Method} & \textbf{Before ($\uparrow$)} & \textbf{After ($\uparrow$)} & \textbf{Eff. ($\downarrow$)} \\
\hline
SAeUron & Baseline & 83.70 & 34.20 & 49.50 \\
        & + \method ($p=1$) & 84.60 & 28.30 & 56.30 \\
        & + \method ($p=2$) & 84.50 & 28.80 & 55.70 \\
\hline
SAEmnesia & Baseline & 97.60 & 57.50 & 40.10 \\
        & + \method ($p=1$) & 91.10 & 56.20 & \textbf{34.90} \\
        & + \method ($p=2$) & 93.20 & 54.60 & 38.60 \\
\hline
\end{tabular}
\end{table}

\subsection{Additional Experiments on UnlearnCanvas.}
\label{sec:appendix_unlearncanvas}

The experiments of~\cref{tab:core_results_saeuron_appendix,tab:core_results_saemnesia_appendix,tab:core_results_gsae_appendix} extend the UnlearnCanvas results of Section~\ref{sec:quantitative_results} by reporting performance for all three padding values $p \in \{0, 1, 2\}$ across the three pipelines. For SAeUron (Table~\ref{tab:core_results_saeuron_appendix}), \method with $p=2$ achieves the highest average (85.80\%), narrowly outperforming $p=1$ (84.35\%), driven by a gain in UA (88.82\% vs.\ 85.37\%), while $p=0$ undershoots the SAeUron baseline (80.88\% vs.\ 82.29\%) due to low UA. For SAEmnesia (Table~\ref{tab:core_results_saemnesia_appendix}), \method with $p=1$ achieves the best average (93.45\%), outperforming both the SAEmnesia baseline (91.51\%) and the $p=2$ variant (91.70\%), with particularly strong CRA (97.97\%); $p=0$ already matches the baseline in average score despite lower UA. For G-SAE (Table~\ref{tab:core_results_gsae_appendix}), \method consistently surpasses the G-SAE baseline (89.94\%) for all padding choices, with $p=1$ yielding the best average (95.33\%) and $p=0$ offering a competitive alternative (95.22\%); $p=2$ lags behind due to a drop in IRA (84.61\%). Across all pipelines, $p=1$ offers the best overall trade-off, achieving top or near-top average scores in every setting while maintaining balanced UA, IRA, and CRA; $p=0$ tends to sacrifice UA, and $p=2$ can degrade IRA, making $p=1$ the most robust default choice.

\begin{table}[!ht]
\setlength{\tabcolsep}{6pt}
\small
\centering
\caption{\textbf{Evaluation metrics (\%) of SAeUron multiplier based pipeline against \method methodology applied on object concept unlearning using the UnlearnCanvas benchmark.} The best result is highlighted in bold.}
\label{tab:core_results_saeuron_appendix}
\begin{tabular}{lccccc}
\hline
\textbf{Method} & \textbf{Padding} & \textbf{UA} (↑) & \textbf{IRA} (↑) & \textbf{CRA} (↑) & \textbf{Avg.} (↑)\\
\hline
SAeUron & -- & 87.16 & 85.57 & 74.14 & 82.29 \\
\method & 0 & 49.12 $\pm$ 4.3 & 95.57 $\pm$ 1.3 & 97.95 $\pm$ 0.9 & 80.88 $\pm$ 1.4 \\
\method & 1 & 85.37 $\pm$ 5.8 & 81.14 $\pm$ 1.4 & 86.55 $\pm$ 1.2 & 84.35 $\pm$ 1.6 \\
\method & 2 & 88.82 $\pm$ 4.1 & 84.57 $\pm$ 1.5 & 84.01 $\pm$ 1.3 & \textbf{85.80} $\pm$ 1.4 \\
\hline
\end{tabular}
\end{table}

\begin{table}[!ht]
\setlength{\tabcolsep}{6pt}
\small
\centering
\caption{\textbf{Evaluation metrics (\%) of SAEmnesia multiplier based pipeline against \method methodology applied on object concept unlearning using the UnlearnCanvas benchmark.} The best result is highlighted in bold.}
\label{tab:core_results_saemnesia_appendix}
\begin{tabular}{lccccc}
\hline
\textbf{Method} & \textbf{Padding} & \textbf{UA} (↑) & \textbf{IRA} (↑) & \textbf{CRA} (↑) & \textbf{Avg.} (↑)\\
\hline
SAEmnesia & -- & 94.65 & 91.39 & 88.48 & 91.51 \\
\method & 0 & 77.84 $\pm$ 2.9 & 95.73 $\pm$ 1.2 & 99.28 $\pm$ 0.4 & 90.95 $\pm$ 1.3 \\
\method & 1 & 91.37 $\pm$ 2.4 & 91.92 $\pm$ 1.4 & 97.97 $\pm$ 0.5 & \textbf{93.45} $\pm$ 1.1 \\
\method & 2 & 93.92 $\pm$ 2.5 & 85.16 $\pm$ 1.3 & 96.02 $\pm$ 0.7 & 91.70 $\pm$ 1.2 \\
\hline
\end{tabular}
\end{table}

\begin{table}[!ht]
\setlength{\tabcolsep}{6pt}
\small
\centering
\caption{\textbf{Evaluation metrics (\%) of G-SAE pipeline against \method methodology applied on object concept unlearning using the UnlearnCanvas benchmark.} The best result is highlighted in bold.}
\label{tab:core_results_gsae_appendix}
\begin{tabular}{lccccc}
\hline
\textbf{Method} & \textbf{Padding} & \textbf{UA} (↑) & \textbf{IRA} (↑) & \textbf{CRA} (↑) & \textbf{Avg.} (↑)\\
\hline
\textsc{G-SAE} & -- & 78.14 & 96.14 & 95.56 & 89.94  \\
\method & 0 & 93.33 $\pm$ 2.3 & 95.39 $\pm$ 1.1 & 96.95 $\pm$ 0.3 & 95.22 $\pm$ 1.0 \\
\method & 1 & 94.02 $\pm$ 2.1 & 96.11 $\pm$ 1.2 & 95.87 $\pm$ 0.4 & \textbf{95.33} $\pm$ 1.1 \\
\method & 2 & 91.72 $\pm$ 2.4 & 84.61 $\pm$ 1.3 & 97.01 $\pm$ 0.3 & 91.11 $\pm$ 1.1 \\
\hline
\end{tabular}
\end{table}

\noindent\textbf{IRA images quality assesment.}
Table~\ref{tab:fid_scores} reports FID scores averaged across 20 UnlearnCanvas objects. The scores are computed on IRA images, evaluating generation quality on concepts unrelated to the unlearned one. Across all pipelines, \method consistently reduces FID compared to the respective baseline, confirming that patch substitution improves image quality for unrelated concepts.

\begin{table}[!ht]
\centering
\caption{\textbf{FID scores ($\downarrow$) on IRA images} for each pipeline. Lower is better; SDXL Turbo denotes the unmodified baseline.}
\label{tab:fid_scores}
\begin{tabular}{llc}
\hline
\textbf{Pipeline} & \textbf{Method} & \textbf{FID ($\downarrow$)} \\
\hline
SAeUron   & Baseline        & 50.56 \\
          & + \method       & \textbf{34.88} \\
\hline
SAEmnesia & Baseline        & 40.82 \\
          & + \method       & \textbf{33.35} \\
\hline
G-SAE     & Baseline        & 39.38 \\
          & + \method       & \textbf{36.19} \\
\hline
SDXL Turbo      & Baseline        & 80.58 \\
          & + \method       & \textbf{59.39} \\
\hline
\end{tabular}
\end{table}

\subsection{Multipliers Ranges.}
\label{sec:appendix_gammas}
In~\cref{tab:gamma_ranges} are reported the value ranges for $\gamma$ employed by SAeUron, SAEmnesia, G-SAE, and SAE on SDXL Turbo respectively. In the case of SAEmnesia, SAeUron and SAE on SDXL Turbo, the ranges were provided by the the authors. for G-SAE, the values of the range are based on a grid search based on the unlearning performances. 
\begin{table}[h]
\centering
\caption{Multiplier $\gamma$ ranges used per method.}
\begin{tabular}{lc}
\toprule
\textbf{Method} & \textbf{$\gamma$ Range} \\
\midrule
SAeUron / SAEmnesia & $\{-1, -5, -10, -15, -20, -25, -30\}$ \\
G-SAE               & $\{-1, -1.5, -2, -3, -4, -5\}$ \\
SAE on SDXL Turbo         & $\{-1, -5, -10, -15, -20, -25, -30, -35, -40\}$ \\
\bottomrule
\end{tabular}
\label{tab:gamma_ranges}
\end{table}
%
%

\subsection{Per-Concept Artifact Rate for SAEmnesia.}
\label{sec:appendix_per_concept_saemnesia}

\begin{figure}[h]
    \centering
    \includegraphics[width=\linewidth]{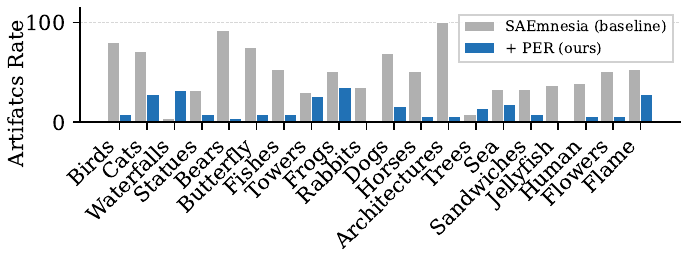}
    \caption{\textbf{Per-concept AR (\%) on UnlearnCanvas for SAEmnesia with and without \method.} Per-concept variation reflects differences in SAE detection quality rather than intrinsic object properties.}
    \label{fig:artifact_comparison_saemnesia}
\end{figure}

Figure~\ref{fig:artifact_comparison_saemnesia} shows the per-concept AR breakdown for SAEmnesia, analogous to Fig.~\ref{fig:infra_class_artifacts} in the main paper for SAeUron. While \method consistently reduces the overall AR from 49.8\% to 15.5\%, the degree of improvement varies across concepts. This variation is not an intrinsic property of the objects themselves, but rather reflects differences in SAE detection quality.

\subsection{Patch Replacement and Activation Manifold.}
\label{sec:appendix_patch_ood}

\begin{figure}[h]
    \centering
    \includegraphics[width=\linewidth]{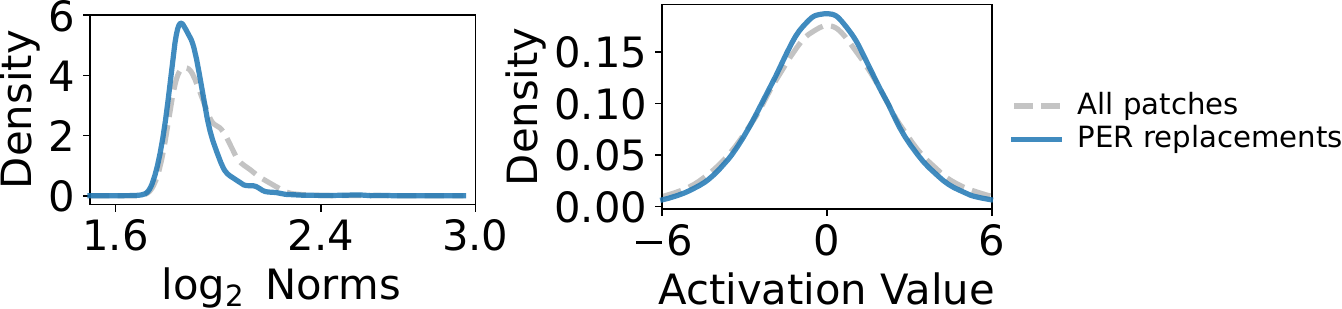}
    \caption{\textbf{Activation distributions of replacement patches vs.\ all patches across concepts.} The distributions are closely aligned, showing how \method keeps activations closer to the learned manifold compared to the multiplier-based steering.}
    \label{fig:patch_replacement_activation}
\end{figure}

A potential concern with \method is that replacing concept-containing patches with patches drawn from the same feature map may itself introduce a distributional shift, since the substituted patches are placed in a spatial context different from their origin. Figure~\ref{fig:patch_replacement_activation} compares the activation distributions of replacement patches against those of all patches in the feature map. The shift introduced by \method is substantially smaller than that caused by multiplier-based steering, where activations are pushed far outside the training distribution (Figure~\ref{fig:sae_distribution} in the main paper), leading to severe visual artifacts.

\end{document}